%% file: root.tex
\documentclass[letterpaper, 10 pt, conference]{ieeeconf}  % Comment this line out if you need a4paper

\IEEEoverridecommandlockouts                              % This command is only needed if
\usepackage{graphics} % for pdf, bitmapped graphics files
\usepackage{epsfig} % for postscript graphics files
\usepackage{times} % assumes new font selection scheme installed
\usepackage{amsmath} % assumes amsmath package installed
\usepackage{amssymb}  % assumes amsmath package installed
\usepackage{lipsum}
\usepackage[ruled,vlined]{algorithm2e}
\usepackage[utf8]{inputenc} %
\usepackage[T1]{fontenc}    %
\usepackage{url}            %
\usepackage{booktabs}       %
\usepackage{amsfonts}       %
\usepackage{nicefrac}       %
\usepackage{microtype}      %
\usepackage[table]{xcolor}  %

\usepackage[
    backend=bibtex,
    style=ieee,
    citestyle=numeric-comp,
    sorting=none,
    sortcites=true,
    maxbibnames=1,
    minbibnames=1,
    giveninits=true,
    doi=false,
    isbn=false,
    url=false,
    eprint=false
]{biblatex}

\AtEveryBibitem{%
    \clearfield{pages}%
    \clearfield{pagetotal}%
    \clearfield{volume}%
    \clearfield{number}%
    \clearfield{issue}%
    \clearfield{month}%
    \clearfield{day}%
    \clearfield{edition}%
    \clearfield{publisher}%
    \clearfield{address}%
    \clearfield{location}%
    \clearfield{organization}%
    \clearfield{institution}%
    \clearfield{school}%
    \clearfield{series}%
    \clearfield{chapter}%
    \clearfield{type}%
    \clearfield{note}%
    \clearfield{addendum}%
    \clearfield{doi}%
    \clearfield{url}%
    \clearfield{urldate}%
    \clearfield{isbn}%
    \clearfield{issn}%
    \clearfield{eprint}%
    \clearfield{eprinttype}%
    \clearfield{eprintclass}%
}

\usepackage{hyperref}
\usepackage[capitalise]{cleveref}
\usepackage{graphicx}
\usepackage{caption}
\usepackage{wrapfig}
\usepackage{tabularx}
\usepackage{cuted}
\usepackage{soul}
\usepackage{mathtools}
\usepackage{multirow}
\usepackage{etoolbox}

\definecolor{algolinegray}{gray}{0.5}

\makeatletter
\patchcmd{\algocf@Vline}
  {\vrule}
  {\color{algolinegray}\vrule\color{black}}
  {}{}

\patchcmd{\algocf@Vsline}
  {\vrule}
  {\color{algolinegray}\vrule\color{black}}
  {}{}

\patchcmd{\algocf@Hlne}
  {\hrule}
  {\color{algolinegray}\hrule}
  {}{}

\title{\LARGE \bf
GUARD: Anomaly-Informed Q-Steering for Generative Robot Policies in Dexterous Manipulation
}

\author{Aryaman Gupta$^{1*}$, Xiaoyue Wu$^{2}$, Rana Soltani Zarrin$^{2}$, Mark Van der Merwe$^{2}$% <-this % stops a space
\thanks{$^{1}$ Stanford University. $^{2}$ Honda Research Institute USA. Email: \texttt{mark\_vandermerwe@honda-ri.com}.}
\thanks{*Work done while interning at Honda Research Institute USA.}}
\begin{document}

\maketitle
\thispagestyle{empty}
\pagestyle{empty}

% \begin{strip}
% \begin{minipage}{\textwidth}\centering
% \vspace{-5em}
% \includegraphics[width=\textwidth]{figs/teaser.pdf}
% \captionof{figure}{teaser}
% \label{fig:teaser}
% % \vspace{-1em}
% \end{minipage}
% \end{strip}

%%%%%%%%%%%%%%%%%%%%%%%%%%%%%%%%%%%%%%%%%%%%%%%%%%%%%%%%%%%%%%%%%%%%%%%%%%%%%%%%
\begin{abstract}

Generative robot policies based on diffusion and flow matching can model complex, multimodal behaviors, but remain susceptible to covariate shift during closed-loop execution. Existing approaches mitigate such failures through additional demonstrations, policy fine-tuning, or test-time guidance, but often require human intervention, engineered rewards, or other task-specific signals. We introduce \textbf{GUARD}, an anomaly-informed Q-steering framework that leverages the policy's training distribution to construct a dense, failure-anticipatory steering signal. GUARD pairs sparse task rewards with dense anomaly scores learned from policy training data and propagates both through a learned Q-function. Its gradients then steer the denoising process toward actions that improve task return while preventing trajectories from entering poorly supported regions, without directly modifying the pretrained policy. Across four Adroit dexterous manipulation tasks, GUARD achieves average success rates of $96.0\%$ and $80.0\%$ under nominal and perturbed dynamics, outperforming the strongest baseline by $21.0$ and $20.0$ percentage points while maintaining the lowest anomaly score. GUARD further demonstrates direct sim-to-real transfer on in-hand screwdriver rotation and reorientation tasks, achieving $40$ and $25$ percentage-point gains in success rate. These results demonstrate that anomaly signals can serve not only for failure monitoring but also to complement sparse task rewards as effective supervision for robust policy steering.

\end{abstract}

%%%%%%%%%%%%%%%%%%%%%%%%%%%%%%%%%%%%%%%%%%%%%%%%%%%%%%%%%%%%%%%%%%%%%%%%%%%%%%%%

\input{sections/1_intro_mark}
\input{sections/2_related_mark}
\input{sections/3_background}
% \input{sections/4_problem}
\input{sections/5_method}
\input{sections/6_experiments}
\input{sections/7_conclusion}

% \input{tables/real_world}

% \addtolength{\textheight}{-12cm}   % This command serves to balance the column lengths
                                  % on the last page of the document manually. It shortens
                                  % the textheight of the last page by a suitable amount.
                                  % This command does not take effect until the next page
                                  % so it should come on the page before the last. Make
                                  % sure that you do not shorten the textheight too much.

%%%%%%%%%%%%%%%%%%%%%%%%%%%%%%%%%%%%%%%%%%%%%%%%%%%%%%%%%%%%%%%%%%%%%%%%%%%%%%%%

%%%%%%%%%%%%%%%%%%%%%%%%%%%%%%%%%%%%%%%%%%%%%%%%%%%%%%%%%%%%%%%%%%%%%%%%%%%%%%%%

%%%%%%%%%%%%%%%%%%%%%%%%%%%%%%%%%%%%%%%%%%%%%%%%%%%%%%%%%%%%%%%%%%%%%%%%%%%%%%%%
% \section*{APPENDIX}

% Appendixes should appear before the acknowledgment.

% \section*{ACKNOWLEDGMENT}

%%%%%%%%%%%%%%%%%%%%%%%%%%%%%%%%%%%%%%%%%%%%%%%%%%%%%%%%%%%%%%%%%%%%%%%%%%%%%%%%

\printbibliography

\end{document}

%% file: sections/1_intro_mark.tex
\section{Introduction}

Generative policies, such as diffusion and flow matching, have emerged as the state-of-the-art for imitation learning in robotics, given their ability to model complex, multimodal action distributions~\cite{chi2025diffusionpolicy,lipman2023flow,black2024pi0}. However, as with other behavior-cloned policies, their performance can degrade under covariate shift: execution errors and environmental variations may drive the system into states poorly represented in the demonstration data, where the learned policy becomes unreliable. This problem is especially pronounced in dexterous manipulation, where complex contact-rich dynamics and long-horizon interactions can amplify small deviations.

\begin{figure}
    \centering
    \includegraphics[width=\linewidth]{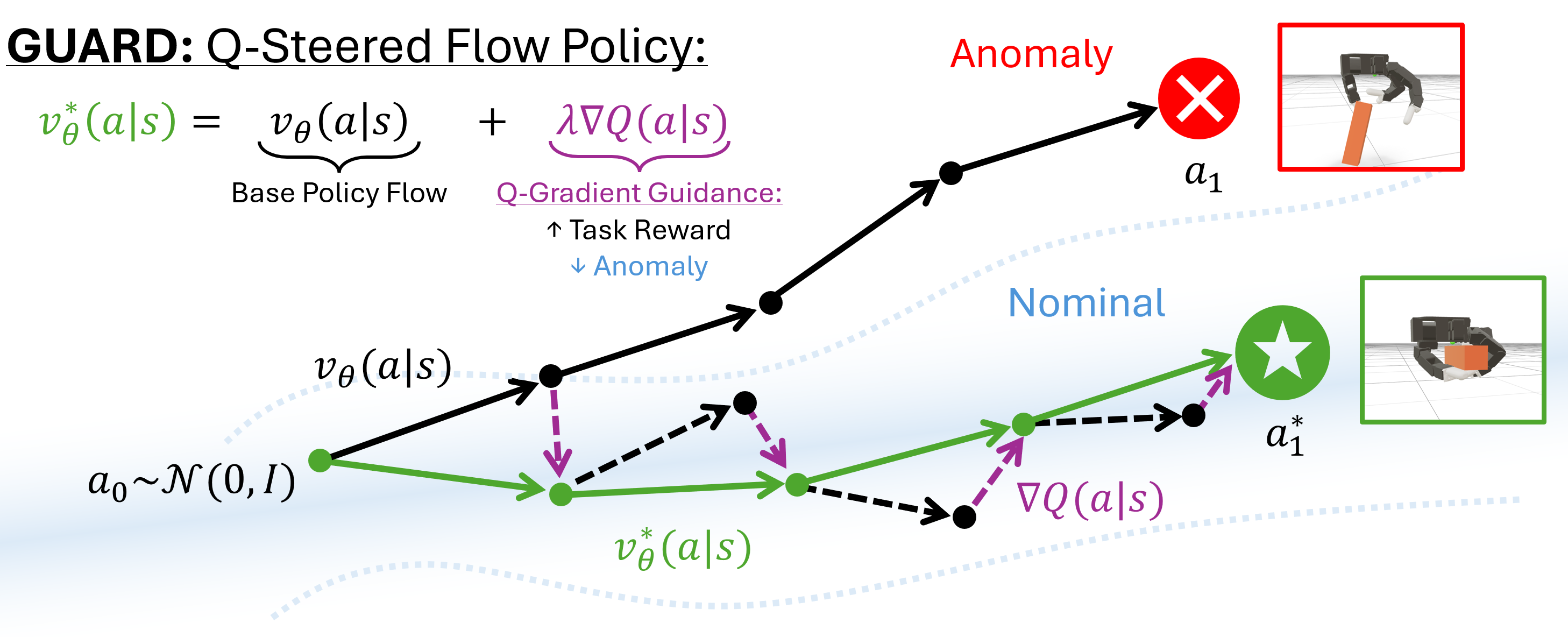}
    \caption{We propose GUARD, a method for steering pretrained generative policies away from failures. GUARD learns a Q-function from sparse task rewards and dense anomaly signals, thus anticipating the risk of entering anomalous regions and the likelihood of task success. The Q-function gradients provide rich guidance to the policy’s denoising process toward successful, well-supported actions.}
    \label{fig:teaser}
\end{figure}

Current methods for addressing covariate shift rely primarily on providing additional signal to adjust policy behavior. Human-in-the-loop methods utilize a supervisor to intervene and provide corrective demonstrations that address deviations as they occur~\cite{ross2011reduction,liu2024model,hoque2024intervengen,pmlr-v164-hoque22a}, but require significant human effort and are difficult to collect for dexterous robot hands. %\delcand{Data augmentation techniques can broaden the coverage of demonstrations but rely on domain knowledge and brittle assumptions~\cite{pmlr-v305-lum25a,pmlr-v229-mandlekar23a}.} 
Methods for post-training generative policies with reinforcement learning can introduce more robust behaviors via trial-and-error~\cite{dppo,reinflow,ankile2024imitation,dsrl,su2026rfs,dicerl} but rely heavily on the quality of the reward signal. Dense rewards provide rich, informative signals to guide policy improvement, but generally require task-specific shaping, additional failure data, or manual failure annotations~\cite{chen2026sarm2,fang2026denserewarddenserewardlearning,ayalew2025progressor}. Sparse task-success rewards are often more scalable, but provide delayed signaling, which can increase the data burden and make adjustments challenging to learn~\cite{andrychowicz2017hindsight,nair2018overcoming}.

% \begin{figure*}[!t]
% \centering
% \includegraphics[width=\textwidth]{figs/method.pdf}
% \caption{Method overview.}
% \label{fig:method}
% \end{figure*}

In this work, we explore how we can construct guidance for a generative policy without relying on human input, brittle assumptions, or engineered signals. Our key insight is to explicitly ``nudge'' the policy back toward its support region rather than relying on laborious task signaling. This prevents our target system from drifting into regions of limited support where our policy is likely to fail. We build on recent work that derives out-of-distribution (OOD) and anomaly scores for robotic behaviors~\cite{fail_detect,fiper,grambow2026anomaly} by constructing a model of the policy's training demonstrations. While previous methods demonstrated the efficacy of anomaly detection for failure monitoring, behavior adjustment is primarily limited to naive retries~\cite{NEURIPS2023_893a5db6} or state-based interpolation~\cite{reichlin2022back,gao2025out}, neither of which can be applied to contact-rich dexterous manipulation.

To translate anomaly detection into effective online adjustment, we propose steering the action-generation process at inference time. Inspired by gradient guidance in generative modeling~\cite{classifier_guidance}, test-time steering preserves the pretrained policy as a behavioral prior while modifying its samples toward desirable actions. To effectively steer, we require an \textit{anticipatory} signal of covariate shift, i.e., grounding future actions to their likelihood of leading to anomalous states. We propose constructing a Q-function trained on offline policy rollouts and online environment interactions to capture the discounted likelihood of future anomalies. The result is a function we can directly utilize to steer a policy to remain near its support. Further, we compose our proposed signal with a sparse task reward, which specifies \emph{what} condition should ultimately be achieved, while the anomaly signal indicates \emph{where} the pretrained policy is likely to remain competent. The resulting value function captures both the task return and the risk of leaving high-support regions, and its action gradient is used to steer each denoising process toward actions that improve the predicted task outcome while remaining close to the trusted region.

% We demonstrate our proposed approach across several simulated and one real task. Our results demonstrate that steering using the anomaly signal, paired with sparse task reward signals, yields robust policies. Additionally, we show that our Q-function can be used to generalize our method to new environments, mitigating errors arising from changes in dynamics and during sim-to-real deployment.

%We show a comprehensive evaluation on contact-rich dexterous manipulation tasks in the Adroit simulation benchmark and through direct sim-to-real transfer on real-world tasks.

\textbf{Contributions.} We introduce \textbf{GUARD}: \underline{G}uidance \underline{U}sing \underline{A}nomaly-info\underline{R}med \underline{D}ense Values, a method that \textbf{(1)} combines sparse task rewards with dense anomaly signals to learn informative Q-functions for steering; and \textbf{(2)} utilizes an offline-to-online Q-function learning mechanism that uses the resulting value gradients to steer frozen flow matching policies toward successful, well-supported actions. We present a comprehensive evaluation on contact-rich dexterous manipulation tasks in the Adroit simulation benchmark and via direct sim-to-real transfer across three real-world tasks.
% comparing against sparse value guidance and latent dynamics-based steering methods on flow matching policies.
% \agnote{1,2 seem similar}

%% file: sections/2_related_mark.tex
\section{Related Works}
\label{sec:related}

\noindent \textbf{Failure Prediction and Anomaly Detection.}
Several recent works study failure prediction through anomaly detection, OOD detection, and uncertainty estimation. FAIL-Detect learns a flow model on the observation distribution from policy training data without requiring failure examples~\cite{fail_detect}, while FIPER combines observation OOD and action-distribution uncertainty to predict failures of generative policies~\cite{fiper}. Sentinel monitors temporal action consistency and task progress~\cite{sentinel}, and other anomaly-based approaches detect deviations from nominal manipulation behavior to trigger failsafe or recovery mechanisms~\cite{grambow2026anomaly,santhanam2025reliable,rolland2026failure}. Policy uncertainty has also been used to improve residual-RL adaptation~\cite{uncertainty_residual_rl}. In contrast, we use policy-input anomaly scores as \emph{dense reward supervision} for learning Q-functions, allowing information about distributional support to propagate through the Bellman return and shape the same value gradients that steer actions toward task success.

\noindent \textbf{Improving Generative Robot Policies.}

The prevalence of generative policy methods has stoked much interest in methods that address failure modes and improve task performance. Applying post-training Reinforcement Learning (RL) is an intuitive choice, and has been demonstrated with success on generative policies~\cite{dppo,reinflow}, however, the generative denoising processes make training unstable and inefficient. Alternative approaches include applying RL in the latent space of the policy~\cite{dsrl,su2026rfs}, adding a residual policy to the generative policy outputs~\cite{ankile2024imitation,su2026rfs,dicerl}, or ranking multiple outputs from the policy with an external value function~\cite{nakamoto2024vgps}. Several works build on gradient guidance for generative modeling~\cite{classifier_guidance} to inject learned Q-function gradients into the denoising process~\cite{qfql,vgd,qgf,qpilots}. We build on this Q-guided sampling machinery but address an orthogonal bottleneck:
\emph{what signal should the Q-function learn from when task supervision is sparse?}

Several other methods have explored moving beyond RL signals for steering. UA-Flow steers flow generation directly using predicted policy uncertainty~\cite{uaflow}. DynaGuide uses an action-conditioned dynamics model to steer predicted outcomes toward desired behaviors~\cite{dynaguide}. Latent Policy Barrier similarly predicts future latent states and guides them toward the expert demonstration distribution~\cite{lpb}. In contrast, our work utilizes a Q-function approach to transform an anomaly signal into a failure anticipatory signal, thereby avoiding dynamics learning. We provide a direct comparison to these methods in our experiments.

%% file: sections/3_background.tex
\section{Background}
\label{sec:background}

\subsection{Reinforcement Learning}
We consider a Markov decision process (MDP)~\cite{sutton1998reinforcement} $\mathcal{M}=(\mathcal{S},\mathcal{A},P,r,\gamma)$, with state space $\mathcal{S}$, action space $\mathcal{A}$, transition dynamics $P$, reward $r$, and discount factor $\gamma\in[0,1)$. A policy $\pi(a\mid s)$ maximizes the expected discounted return $J(\pi)=\mathbb{E}_{\pi,P}\left[\sum_{t=0}^{\infty} \gamma^t r(s_t, a_t)\right]$. Its action-value function is $Q^\pi(s,a)=\mathbb{E}_{\pi,P}[\sum_{t=0}^{\infty}\gamma^t r(s_t,a_t)\mid s_0=s,a_0=a]$. Given transitions $(s,a,r,s')\sim\mathcal{D}$, a Q-function $Q_\phi$ can be learned through temporal-difference learning by minimizing $\mathcal{L}_{Q}(\phi)=\mathbb{E}_{\mathcal{D}}[(Q_\phi(s,a)-r-\gamma Q_{\bar\phi}(s',a'))^2]$, where $a'\sim\pi(\cdot\mid s')$ and $Q_{\bar\phi}$ denotes a target Q-function. Since \(Q^\pi(s,a)\) depends explicitly on the action, its gradient \(\nabla_a Q^\pi(s,a)\) provides a local direction toward actions with a higher predicted return.

\subsection{Flow Matching}
Flow matching~\cite{lipman2023flow,liu2023rectifiedflow} policies are parameterized by a state-conditioned, time-dependent velocity field $v_\theta(a,k;s)$ that transports samples from a Gaussian noise distribution $p_0=\mathcal{N}(0,I)$ to the policy distribution $\pi_\theta(a\mid s)$ via an ODE:
$
\mathrm{d}a_k = v_\theta(a_k,k;s)\mathrm{d}k, \quad a_0 \sim p_0.
$
Given dataset state-action pairs $(s,a_1)\sim\mathcal{D}$, we define linear interpolants $a_k=(1-k)a_0+ka_1$ between noise and data actions and train the velocity field via flow matching:
$\mathcal{L}_{\mathrm{FM}}(\theta) =
\mathbb{E}_{k \sim \mathcal{U}[0,1], a_0 \sim p_0, (s,a_1) \sim \mathcal{D}}
\left[
\left|
v_\theta(a_k,k;s) - (a_1-a_0)
\right|_2^2
\right]$.
At inference time, actions are generated by integrating the ODE from $k=0$ to $1$, yielding a sample $a_1\sim\pi_\theta(a\mid s)$. 
% \agnote{if possible prevent using hat since it is being used for 1-step denoising}

% \subsection{Policy Uncertainty and Anomaly Detection}
% Epistemic uncertainty captures uncertainty arising from limited coverage of the training distribution and can therefore indicate when a policy is operating in poorly supported OOD regions. We denote by $U(s)$ an anomaly score computed from the policy input, where larger values indicate greater deviation from the distribution represented in the demonstration data. Unlike sparse task reward, which specifies whether the desired behavior is ultimately achieved, $U(s)$ provides a dense, task-agnostic signal of policy support.
% We leverage this complementary signal to construct the reward used for learning our steering Q-functions.

%% file: sections/5_method.tex
\section{\textbf{GUARD}: \underline{G}uidance \underline{U}sing \underline{A}nomaly-info\underline{R}med \underline{D}ense Values}
\label{sec:method}

Our method consists of three components: (i) anomaly signals-informed dense reward construction, (ii) Q-guided action denoising, and (iii) offline-to-online Q-function adaptation. The overall procedure is summarized in Alg.~\ref{alg:guide_training}.

\subsection{Anomaly Signals-Informed Dense Rewards}
\label{sec:dense_reward}

Let $\mathcal{D}_{\mathrm{demo}}$ denote the demonstration data used to train the frozen flow policy $\pi_\theta$. We fit an anomaly detector $U(s)$ on the policy inputs in $\mathcal{D}_{\mathrm{demo}}$, where larger values indicate observations that are less supported by the policy's training distribution. Since $U$ is learned solely from the policy training data, it requires neither failure examples nor task-specific failure annotations. Our method naturally admits many available anomaly metrics.
% in our experiments, we primarily utilize the logpZO score~\cite{fail_detect} by fitting a separate flow model to observations from $\mathcal{D}_{\mathrm{demo}}$ and using the magnitude of the latent corresponding to an observation as its anomaly score.
In our experiments, we primarily use the logpZO score~\cite{fail_detect} by fitting a separate flow model to the observations in $\mathcal{D}_{\mathrm{demo}}$ that transports each observation to a Gaussian latent $z$, and take the squared latent norm $\lVert z\rVert_2^2$ as its continuous anomaly score (large when an observation maps far from in-distribution regions).

% We use $U$ to provide dense transition-level supervision. Since action $a_t$ induces the transition to $s_{t+1}$, we associate its anomaly reward with the resulting state. We first collect an offline rollout buffer $\mathcal{D}_{\mathrm{off}}$ using the frozen flow policy and compute the reward-normalization statistics
% $\Xi_{\mathrm{off}}
% \coloneqq
% \{\mu_{\log U}^{\mathrm{off}},\sigma_{\log U}^{\mathrm{off}},
% \mu_{\mathrm{task}}^{\mathrm{off}},\sigma_{\mathrm{task}}^{\mathrm{off}}\}$
% once over this buffer. We keep $\Xi_{\mathrm{off}}$ fixed throughout offline and online learning and define the dense anomaly reward using log-transformed values as

We assign each state-action pair $(s_t,a_t)$ the anomaly score of $s_{t+1}$, i.e., $U(s_{t+1})$. Using an offline buffer $\mathcal{D}_{\mathrm{off}}$ collected with the frozen flow policy, we compute and freeze reward-normalization statistics
$\Xi_{\mathrm{off}}
\coloneqq
\{\mu_{\log U}^{\mathrm{off}},\sigma_{\log U}^{\mathrm{off}},
\mu_{\mathrm{task}}^{\mathrm{off}},\sigma_{\mathrm{task}}^{\mathrm{off}}\}$.
The dense anomaly reward is then defined as
\begin{equation}
    r_t^{U}
    =
    -\frac{
        \log U(s_{t+1})-\mu_{\log U}^{\mathrm{off}}
    }{
        \sigma_{\log U}^{\mathrm{off}}
    }
    \label{eq:u_reward}
\end{equation}
Thus, transitions toward states better supported by the policy's training distribution receive a larger reward, while transitions toward anomalous states are penalized.

The available task supervision is a sparse binary signal, $r_t^{\mathrm{task}}$. To place both rewards on comparable scales, we independently standardize the task reward using the fixed statistics in $\Xi_{\mathrm{off}}$,
\begin{equation}
    \bar r_t^{\mathrm{task}}
    =
    \frac{
        r_t^{\mathrm{task}}-\mu_{\mathrm{task}}^{\mathrm{off}}
    }{
        \sigma_{\mathrm{task}}^{\mathrm{off}}
    },
    \label{eq:task_reward_norm}
\end{equation}
and define the reward used for Q-learning as their equal-weight sum,
\begin{equation}
    \tilde r_t
    =
    r_t^{U}
    +
    \bar r_t^{\mathrm{task}}.
    \label{eq:aug_reward}
\end{equation}
Independent standardization places both signals on comparable scales without introducing a tunable trade-off coefficient.

The two rewards provide complementary supervision: the sparse task reward specifies \emph{what} the robot should ultimately accomplish, while the dense anomaly reward indicates \emph{where} the pretrained policy is well supported. Crucially, we do not directly steer actions to minimize instantaneous anomaly. Instead, Eq.~\eqref{eq:aug_reward} enters the Bellman return, allowing the Q-function to propagate both task success and distributional support over future trajectories.

\subsection{Q-Guided Action Denoising}
\label{sec:q_guidance}

We learn an ensemble of $N$ independently initialized Q-functions,
$\mathcal{Q}_{\phi}=\{Q_{\phi_i}\}_{i=1}^{N}$. To reduce exploitation of Q-function extrapolation errors during gradient-based guidance, we follow prior Q-guided methods~\cite{qpilots} and construct the pessimistic ensemble estimate
\begin{equation}
    L(s,a;\mathcal{Q}_{\phi})
    =
    \frac{1}{N}\sum_{i=1}^{N} Q_{\phi_i}(s,a)
    -
    \rho\,
    \underset{i}{\operatorname{std}}\,Q_{\phi_i}(s,a),
    \label{eq:pessimistic_q}
\end{equation}
where $\rho\geq0$ controls the degree of pessimism. The ensemble disagreement in Eq.~\eqref{eq:pessimistic_q} is distinct from the anomaly score $U(s)$: $U(s)$ measures support of the \emph{policy input} under the training distribution and is used for reward construction, whereas Q-ensemble disagreement provides conservative value estimates for guidance.

Intermediate flow variables are noisy and generally lie outside the clean-action distribution on which the Q-functions are trained. Following prior works~\cite{qgf,edp}, at each denoising step $k\in[0,1]$, we estimate the corresponding clean action with a single step over the remaining flow horizon,
\begin{equation}
    \hat a_k
    =
    a_k
    +
    \left(1-k\right)
    v_\theta\!\left(a_k,k;s\right),
    \label{eq:one_step_denoise}
\end{equation}
query the ensemble at $\hat a_k$ and compute
$g_k=\nabla_{\hat a_k}L(s,\hat a_k;\mathcal{Q}_{\phi})$.
Following QGF~\cite{qgf}, we approximate
$\partial\hat a_k/\partial a_k\approx I$, avoiding differentiation through the flow velocity. With denoising step size $\Delta$, the guided update is
\begin{equation}
    a_{k+\Delta}
    =
    a_k
    +
    \Delta
    \left[
        v_\theta(a_k,k;s)
        +
        \lambda
        \nabla_{\hat a_k}
        L(s,\hat a_k;\mathcal{Q}_{\phi})
    \right].
    \label{eq:guided_flow}
\end{equation}
We denote the action distribution induced by iterating Eq.~\eqref{eq:guided_flow} as
$\pi_{\mathcal{Q}_{\phi}}(a\mid s)$. The frozen flow policy remains the behavioral prior, while the Q-gradient biases denoising toward actions with high future anomaly signal-augmented task return. Here $\lambda$ is the guidance weight, and we discuss details about the selection of $\lambda$ in Sec .~\ref {sec:experiments}.

 % Unlike directly minimizing the instantaneous anomaly score, this guidance accounts for how an action affects both future task success and the likelihood of entering poorly supported states.

\subsection{Offline-to-Online Q-Function Adaptation}
\label{sec:online_q}

The Q-functions used for guidance should evaluate the policy that is ultimately executed, i.e., the steered policy. We therefore first estimate values under the base policy with offline initialization and subsequently adapt them to the evolving steered policy using online experience.

\paragraph{Offline initialization}
\label{sec:offline_qtraining}
We collect an offline buffer $\mathcal{D}_{\mathrm{off}}$ by rolling out the frozen policy $\pi_\theta$ with action noise $\sigma_a$, recording each transition together with its subsequent behavior action $a'_{\mathrm{BC}}$. Each independently initialized ensemble head uses its own exponentially averaged target Q-function $\bar Q_i$. We perform behavior-policy fitted Q evaluation (FQE) using
\begin{equation}
    y_i^{\mathrm{off}}
    =
    \tilde r
    +
    \gamma(1-d)
    \bar Q_i(s',a'_{\mathrm{BC}}),
    \label{eq:offline_td}
\end{equation}
where $d$ is the terminal indicator. Each $Q_{\phi_i}$ is regressed toward its corresponding target, while its target parameters are updated as
$\bar\phi_i\leftarrow\tau\phi_i+(1-\tau)\bar\phi_i$.
Using per-head targets during offline initialization preserves ensemble diversity while estimating the anomaly-augmented return in regions covered by the behavior policy.

\paragraph{Online adaptation}
\label{sec:online_qtraining}
Once Q-guidance is applied, the executed policy
$\pi_{\mathcal{Q}_{\phi}}$
differs from the behavior policy used for offline initialization. A Q-function evaluated only on the original behavior distribution can therefore become mismatched to the policy it is used to guide. We address this by alternating guided interaction with FQE updates under the resulting policy.

At each online iteration, we collect fresh trajectories using
$a\sim\pi_{\mathcal{Q}_{\phi}}(\cdot\mid s)$
and append them to the unified replay buffer
$\mathcal{B}=\mathcal{D}_{\mathrm{off}}\cup\mathcal{D}_{\mathrm{on}}$.
All online transitions are assigned rewards using Eq.~\eqref{eq:aug_reward} with the \emph{fixed} normalization statistics $\Xi_{\mathrm{off}}$.

For a sampled transition $(s,a,\tilde r,s',d)$, we regenerate a fresh bootstrap action using the \emph{guided policy} induced by the EMA target ensemble,
$a'\sim\pi_{\bar{\mathcal{Q}}}(\cdot\mid s')$,
and construct the pessimistic target
\begin{equation}
    y
    =
    \tilde r
    +
    \gamma(1-d)
    L(s',a';\bar{\mathcal{Q}}).
    \label{eq:online_td}
\end{equation}
Each ensemble head is regressed toward this shared target,
$\mathbb{E}_{\mathcal{B}}
[(Q_{\phi_i}(s,a)-y)^2]$,
while its target network is updated by Polyak averaging. Using the slowly evolving target ensemble to generate both $a'$ and its bootstrap value decouples action generation from instantaneous Q-function updates and stabilizes the guidance-value feedback loop. Repeated interaction and evaluation progressively adapt the Q-functions toward the steered policy, so that $Q(s,a)$ estimates the long-horizon return obtained after taking $a$ and subsequently following that guided policy. % Its action gradient can therefore guide denoising toward locally higher-return actions while jointly accounting for both task success and distributional support.
The resulting policy $\pi_{\mathcal{Q}_{\phi}}(a\mid s)$ uses the ensemble of online-adapted Q-functions to guide denoising towards high-return, high-support actions. Refer Alg.~\ref{alg:guide_training} for the full training procedure.

\input{sections/algo}

% We keep the automatically calibrated guidance scale $\lambda$ fixed throughout online adaptation. As the Q-functions improve, we empirically observe the gradient magnitude to decrease; with fixed $\lambda$, this naturally attenuates the magnitude of the steering correction while preserving its value-improving direction, providing meaningful guidance without increasingly aggressive perturbations of the base flow.

%% file: sections/algo.tex
\begin{algorithm}
\SetAlgoVlined
\DontPrintSemicolon
\newcommand{\graykw}[1]{\textcolor{gray}{#1}}
%\SetKwSty{graykw}
%\SetKwSty{color{gray}}
\SetKwInOut{Input}{Require}
\small

\Input{Frozen flow policy $\pi_\theta$; anomaly detector $U$;
ensemble size $N$; pessimism $\rho$; guidance weight $\lambda$, action noise $\sigma_a$;
discount $\gamma$; Polyak rate $\tau$}

Collect $\mathcal{D}_{\mathrm{off}}$ by rolling out $\pi_\theta$
with action noise $\sigma_a$, storing
$(s,a,s',a'_{\mathrm{BC}},r^{\mathrm{task}},U(s'),d)$\;

Compute reward statistics $\Xi_{\mathrm{off}}$ from
$\mathcal{D}_{\mathrm{off}}$ and freeze them\;

$\tilde r \coloneqq
\textsc{Reward}(r^{\mathrm{task}},U(s');\Xi_{\mathrm{off}})$
using~\eqref{eq:aug_reward}; \quad
$\mathcal{B}\leftarrow\mathcal{D}_{\mathrm{off}}$\;

Initialize independent
$\mathcal{Q}_{\phi}=\{Q_{\phi_i}\}_{i=1}^{N}$ and
$\bar Q_i\leftarrow Q_{\phi_i},\ \forall i$\;

$L(s,a;\mathcal{Q})
\coloneqq
\operatorname{mean}_{i}Q_i(s,a)
-\rho\,\operatorname{std}_{i}Q_i(s,a)$\;

$\pi_{\mathcal{Q}}(\cdot\mid s)
\coloneqq$ Q-guided flow policy defined by~\eqref{eq:guided_flow}\;

\BlankLine
\textcolor{gray}{\textit{Phase-I: Offline Q-function initialization}}\;

\For{$g=1,\ldots,G_{\mathrm{off}}$}{
    Sample
    $(s,a,s',a'_{\mathrm{BC}},r^{\mathrm{task}},U(s'),d)
    \sim\mathcal{D}_{\mathrm{off}}$\;

    $\tilde r\leftarrow
    \textsc{Reward}(r^{\mathrm{task}},U(s');\Xi_{\mathrm{off}})$\;

    \For{$i=1,\ldots,N$}{
        $y_i^{\mathrm{off}}\leftarrow
        \tilde r+
        \gamma(1-d)\bar Q_i(s',a'_{\mathrm{BC}})$\;

        Update $Q_{\phi_i}$ toward $y_i^{\mathrm{off}}$; \quad 
        $\bar\phi_i\leftarrow
        \tau\phi_i+(1-\tau)\bar\phi_i$\;
    }
}

\BlankLine
\textcolor{gray}{\textit{Phase-II: Online Q-function adaptation}}\;

\For{$n=1,\ldots,N_{\max}$}{

    Collect fresh transitions using
    $\pi_{\mathcal{Q}_{\phi}}$
    with action noise $\sigma_a$ and append them to $\mathcal{B}$\;

    \For{$g=1,\ldots,G$}{
        Sample
        $(s,a,s',r^{\mathrm{task}},U(s'),d)\sim\mathcal{B}$\;

        $\tilde r\leftarrow
        \textsc{Reward}(r^{\mathrm{task}},U(s');\Xi_{\mathrm{off}})$; \enspace
        $a'\sim
        \pi_{\bar{\mathcal{Q}}}(\cdot\mid s')$\;

        $y\leftarrow
        \tilde r+
        \gamma(1-d)
        L(s',a';\bar{\mathcal{Q}})$\;

        \For{$i=1,\ldots,N$}{
            Update $Q_{\phi_i}$ toward $y$; \enspace
            $\bar\phi_i\leftarrow
            \tau\phi_i+(1-\tau)\bar\phi_i$\;
        }
    }
}

\Return{$\mathcal{Q}_{\phi}$}

\caption{GUARD Training.
}
\label{alg:guide_training}
\end{algorithm}

%% file: sections/6_experiments.tex
\vspace{-1em}
\section{Experiments}
\label{sec:experiments}

\input{tables/adroit_state_4}

We evaluate GUARD on the D4RL Adroit dexterous manipulation benchmark~\cite{rajeswaran2017learning} in simulation and further demonstrate its sim-to-real transfer capabilities across three tasks using the Allegro V5 Plus multi-fingered robot hand. Through these experiments, we aim to answer the following research questions (RQs):

\begin{enumerate}
\item \textbf{RQ1}: Can GUARD reliably steer the policy toward successful task completion?

\item \textbf{RQ2}: Does GUARD keep the system near high-support, in-distribution regions?

\item \textbf{RQ3}: Is GUARD more robust under distribution shift, including perturbed dynamics and sim-to-real transfer?

\end{enumerate}

\subsection{Adroit Simulation Benchmark}

%\agnote{add tasks ss if we have space}
We run experiments on all four Adroit manipulation tasks: \emph{Relocate}, \emph{Pen}, \emph{Hammer}, and \emph{Door}~\cite{rajeswaran2017learning}.
% We also evaluate on the same tasks using vision-input policies, rolling out an RL expert~\cite{xu2024drm} to generate 300 episodes for training the BC policies. 
 We train base flow-matching policies over 200 episodes, sampled equally from the \emph{expert} and \emph{cloned} subsets. The policy uses a four-layer MLP with 512 hidden units, GELU activations, layer normalization, and $K{=}10$ Euler denoising steps. GUARD uses an ensemble of five Q-functions with pessimism $\rho{=}0.5$, trained for $40$k offline FQE updates on $800$ episodes with batch size $256$, learning rate $3{\times}10^{-4}$, discount $\gamma{=}0.99$, EMA target rate $\tau{=}0.005$, followed by $150$ iterations of online adaptation collecting $24$ episodes each iteration. Each episode includes $200$ steps, and the sparse task reward used for Adroit tasks is $-1$ everywhere and $0$ on successful states, aligning with the commonly used choice in offline RL. We calibrate $\lambda$ once after offline initialization by matching the average gradient scale to the average flow-velocity scale,
\(\lambda = \tfrac{\mathbb{E} \left[\lVert v_\theta\rVert_2\right]}
{\mathbb{E} \left[\lVert\nabla_{\hat a_k}
L(s,\hat a_k;\mathcal{Q}_{\phi})\rVert_2\right]}\)
where the expectations are estimated using the offline-initialized ensemble and then hold $\lambda$ fixed throughout the online adaptation and deployment.

\paragraph{Baselines}
We compare GUARD against: (i) \textbf{BC}, the unsteered flow-matching policy; (ii) \textbf{LPB}~\cite{lpb} and (iii) \textbf{DynaGuide}~\cite{dynaguide}, which use a learned dynamics model to steer predicted future states toward in-distribution regions and desired task outcomes respectively; (iv) \textbf{UA-Flow}~\cite{uaflow}, which guides flow sampling toward actions with lower predicted uncertainty; (v) \textbf{Sparse-Q}, Q-guided denoising using only the sparse task success reward; and (vi) \textbf{U-SA}, which directly guides denoising using action gradient of joint state-action anomaly scores $U(s,a)$. Every baseline is trained using its prescribed methodology, with an equal data budget (both offline and online) to GUARD, for fair comparison. UA-Flow and U-SA are trained only with $\mathcal{D}_{\mathrm{demo}}$, as they rely only on the training demonstrations.

\paragraph{Results}

In Table~\ref{tab:adroit_state}, we evaluate $100$ episodes per task in each of the nominal and perturbed dynamics settings. In the latter setting, we add Gaussian noise to the actions to investigate GUARD under distribution shift.

\begin{figure*}[!t]
\vspace{0.5em}
\centering
\includegraphics[width=\textwidth]{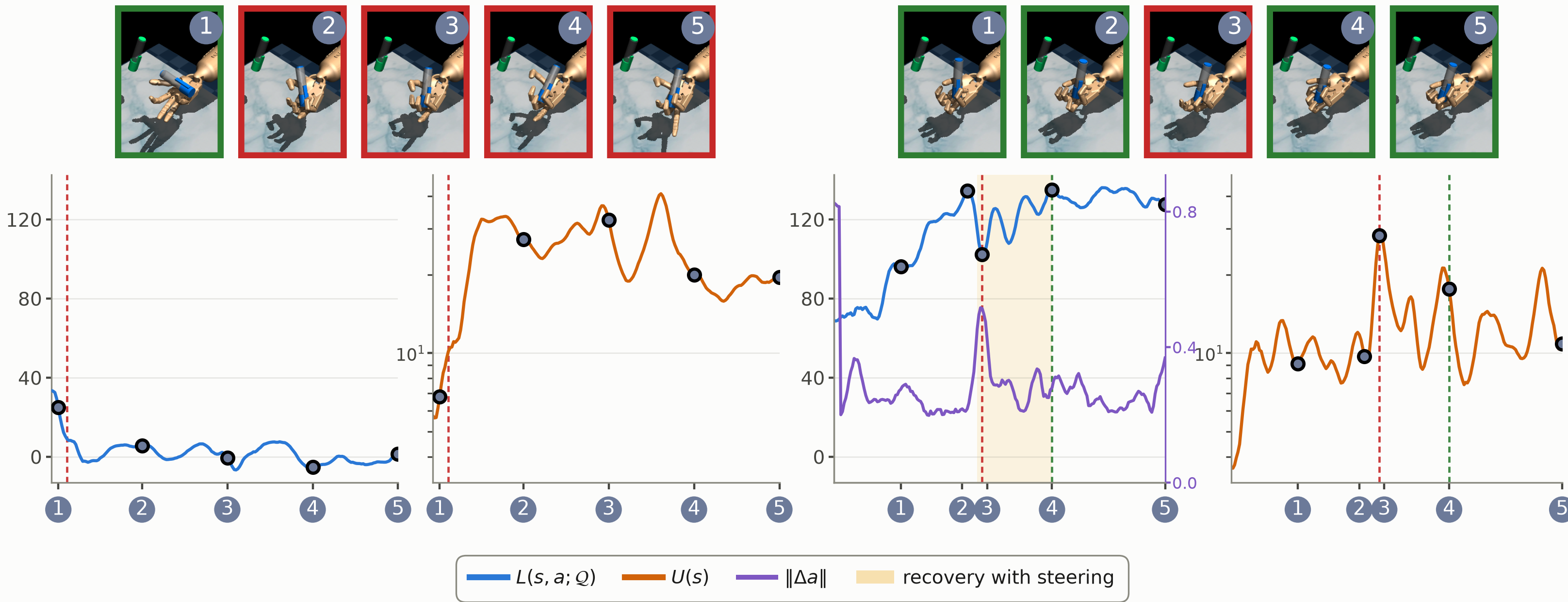}
\caption{A qualitative visualization of steering with GUARD in the Adroit \emph{Pen} task. \textbf{(Left)} unsteered BC policy forms a loose grasp of the pen and fails to properly reorient it, as indicated by low Q-values $L(s,a;\mathcal{Q})$ and high state-anomaly scores $U(s)$. \textbf{(Right)} At the moment some fingers lose contact, steering provides a larger action correction $\|\Delta a\|$ folding the fingers back in to maintain a firm grasp and reorient the pen to the target orientation.}
\label{fig:qual_pen}
\end{figure*}

% As shown in Table~\ref{tab:adroit_state}, when evaluated over $100$ episodes per task i
In the nominal setting, GUARD achieves the strongest overall task performance, reaching an average success rate (SR) of $0.96$. This corresponds to an average gain of $36.0$ percentage points (pp) over the BC policy and a gain of $21.0$ pp over the next-best baseline, DynaGuide. Fig.~\ref{fig:qual_pen} shows how GUARD successfully steers an originally failed trajectory under the BC policy for the \emph{Pen} task starting from the same initial state. These results establish for \textbf{RQ1} that \emph{\ul{GUARD can effectively steer generative policies toward task success and outperform existing steering strategies}}.

GUARD also consistently keeps the system much closer to in-distribution regions, achieving the lowest state-anomaly scores (AS) on successful episodes for each method across all tasks in the nominal setting. Here, the AS is evaluated using random network distillation (RND)~\cite{burda2018exploration, rnd}, a neutral metric for fair comparison across all methods, using a fixed random target network $g$ and a predictor $h$ trained to regress $g$ on $\mathcal{D}_{\mathrm{demo}}$ inputs, calculating the anomaly score as $\mathrm{RND}(s)=\lVert h(s)-g(s)\rVert_2^2$. These results answer \textbf{RQ2}: \emph{\ul{GUARD realizes substantially stronger support under the policy training distribution}}.

\begin{figure*}[htbp]
\vspace{0.5em}
\centering
\includegraphics[width=\textwidth]{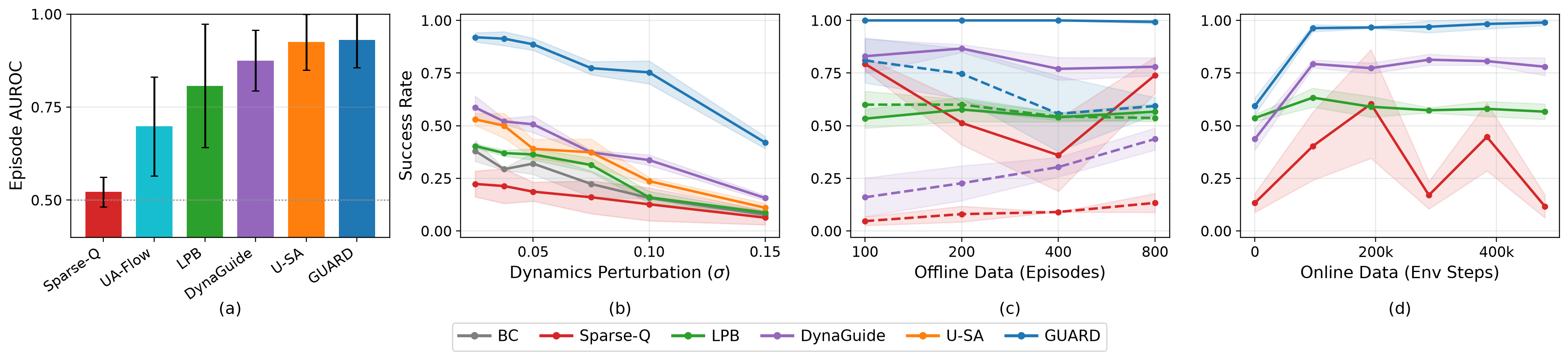}
\caption{\textbf{(a)} Failure-detection quality on \emph{Door}: episode-level AUROC
(max-aggregated per-step score) using each method's steering signal. \textbf{(b)} Robustness on \emph{Door} to test-time dynamics perturbations of
magnitude $\sigma_d$ (action noise). \textbf{(c)}  Offline data efficiency on \emph{Hammer}: success rate vs.\ offline dataset
size; \emph{dashed} = offline-only (no online adaptation), \emph{solid} =
offline\,+\,online training. \textbf{(d)}  Online data efficiency on \emph{Hammer}: success rate vs.\
online interaction (env steps) at the 800-episode offline budget. Overall, \textbf{GUARD} leads across failure detection, dynamics robustness, and data efficiency.}
\label{fig:adroit_ablations}
\end{figure*}

Finally, we compute both SR and AS in the perturbed-dynamics setting. GUARD achieves an average SR of $0.80$ in the shifted environment, a $39.0$ and $20.0$ pp gain over the BC policy and the next-best baseline, again Dynaguide, and the lowest AS  score on average across all tasks. Fig.~\ref{fig:adroit_ablations}(b) shows our method maintains the highest performance with increasing perturbation levels as shown on the \emph{Door} task. Thus, answering \textbf{RQ3}, \emph{\ul{GUARD demonstrates more robust performance across both tests under distribution shift.}}

Our results indicate the efficacy of GUARD for driving task improvement. The performance gains over U-SA (nominal: $+24.0$ pp; perturbed: $+28.0$ pp) indicate the importance of grounding the signal in actions with the Q-function. Gains over the Sparse-Q (nominal: $+30.0$ pp; perturbed: $+30.0$ pp) show the value of dense anomaly-signaling. While LPB and DynaGuide utilize a failure-anticipation mechanism by rolling out the predicted dynamics, we find that GUARD outperforms both, suggesting that the dynamics of the contact-rich tasks under test may be challenging to capture accurately. Finally, UA-Flow relies only on the policy uncertainty, which provides a weaker steering signal than GUARD.

\paragraph{Failure Signal Alignment}
We compare each method's failure-detection performance using the AUROC metric aggregated over the episode. Fig.~\ref{fig:adroit_ablations}(a) shows that GUARD achieves the highest AUROC, demonstrating its ability to reliably identify failures, complementing its steering performance. 

\paragraph{Training Data Efficiency}
\textsc{GUARD} is highly data-efficient in both the offline and online phases. The Q-function reaches near-perfect steered success with only a modest buffer of offline data paired with online training and stays saturated as the buffer grows, whereas the strongest baselines (DynaGuide, LPB) plateau well below it (Fig.~\ref{fig:adroit_ablations}(c)). The gap between the offline-only (dashed) and offline\,+\,online (solid) curves shows that offline training alone is insufficient and that online adaptation supplies most of the leverage for steering. Online, \textsc{GUARD} quickly improves its signaling, climbing to $\approx\!0.9$ within the first $\sim\!10^{5}$ environment steps and saturating near $1.0$, while the baselines improve more slowly and to a lower ceiling  (Fig.~\ref{fig:adroit_ablations}(d)). Thus, \textsc{GUARD} achieves the highest performance with little offline data and some online interactions, a favorable trade-off for a scalable real-world deployment.

\paragraph{Ablation on Anomaly Signal}

We ablate the choice of state-anomaly score $U(s)$ by swapping in alternative OOD signals in GUARD, such as input reconstruction error (Recon-AE)~\cite{grambow2026anomaly}, RND~\cite{rnd}, and learned dynamics-based metrics such as prediction error (Dyn-Err) and ensemble disagreement (Dyn-Ens)~\cite{haider2023out}, as shown in Table~\ref{tab:anomaly_ablation_door}. The state-density logpZO metric we adopt is both the strongest and most stable, while Recon-AE, RND, and Dyn-Ens also remain competitive. In contrast, Dyn-Err collapses, indicating that prediction error conflates model inaccuracy with genuine anomaly and is a poor proxy for how far a state has drifted OOD. This ablation shows that steering benefits from a \emph{state-distributional} notion of anomaly, and that logpZO is a robust default choice.

\input{tables/anomaly_ablation}

\subsection{Real-World Experiments}

We evaluate on three dexterous tasks with a 16-DoF Allegro V5 Plus hand: (1) screwdriver \emph{in-hand rotation}, requiring continuous rotations throughout the episode without dropping, and screwdriver \emph{in-hand reorientation} in (2) \emph{palm-down} and (3) \emph{palm-up} configurations, requiring the screwdriver to be reoriented vertically below and above the palm, respectively, from a horizontal initial pose, just using its fingers in the fixed-wrist hand, and holding it there for $\geq 5$ sec. The BC policies are cloned from $64$, $256$, and $4096$ demonstrations, respectively, for each task, each $40$ sec long. These demonstrations are generated for each task using an RL expert policy trained with PPO~\cite{schulman2017proximalpolicyoptimizationalgorithms} using self-defined dense rewards in the \texttt{mjlab} simulation, with domain randomization of dynamics and observations. The BC policy is a conditional 1D U-Net (channels $256/512/1024$, kernel $5$) taking $41$ dim observations containing joint positions, previously commanded joint positions, and object pose in the palm frame for all tasks. It outputs $16$ dim joint-position deltas of chunk length $4$ with an execution horizon of $2$. It is trained for $100$ epochs with batch size $256$, learning rate $10^{-4}$, cosine schedule, and EMA decay $0.9999$, with $K{=}10$ Euler steps at inference.
GUARD uses an ensemble of five Q-functions with pessimism $\rho{=}0.5$, trained for $200$k offline FQE updates on $2000$ policy rollouts with batch size $2048$, learning rate $10^{-4}$, discount $\gamma{=}0.95$, EMA target rate $\tau{=}0.005$, followed by $\sim 40$ iterations of online adaptation collecting $128$ episodes in each iteration. The sparse task reward is 0 everywhere and 1 on successful states. We keep dynamics and observation domain randomization enabled to train GUARD under the same noise distribution as seen by the BC policy. Note that GUARD is trained \textbf{entirely in simulation}, for both the offline and online stages, and then evaluated directly in the real world to assess its sim-to-real capabilities.

\input{tables/real_world}

\begin{figure*}[!t]
\vspace{0.5em}
\centering
\includegraphics[width=\textwidth]{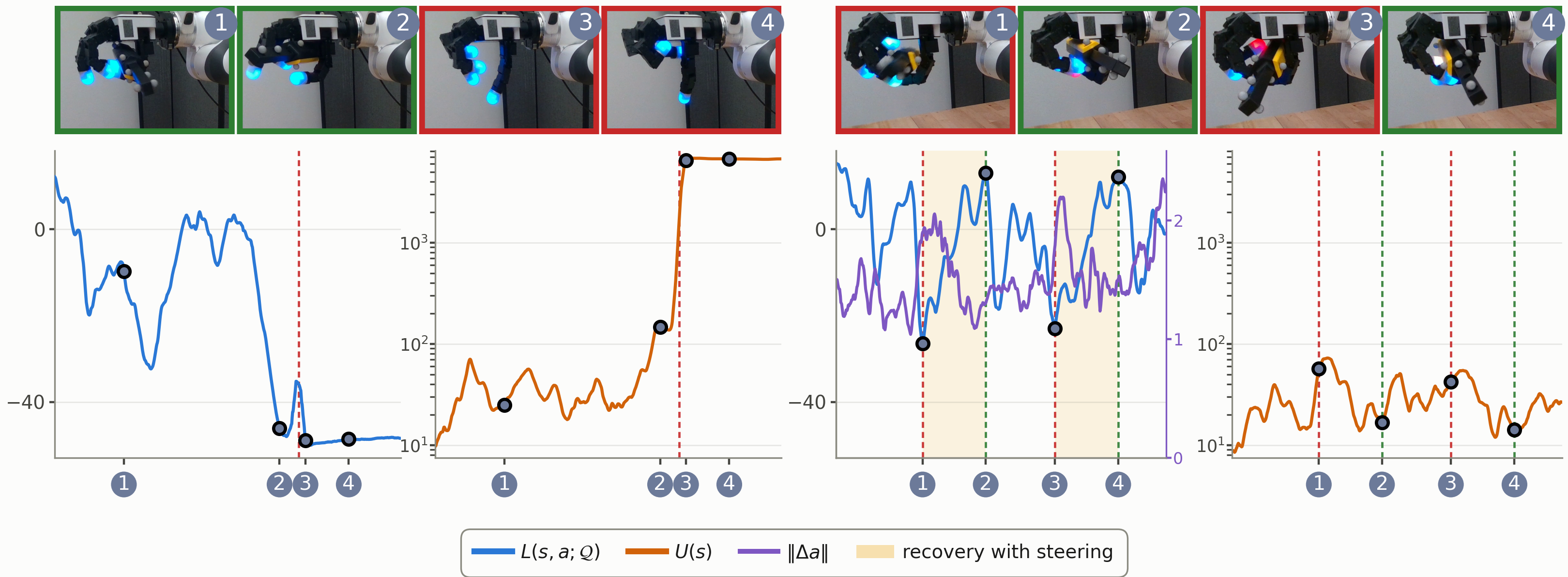}
\caption{A qualitative visualization of steering with GUARD in the screwdriver \emph{in-hand-rotation} task. \textbf{(Left)} unsteered BC policy drops the screwdriver, indicated by a large, sudden drop in the Q-value $L(s,a;\mathcal{Q})$ and a jump in state-anomaly score $U(s)$. \textbf{(Right)} Whenever the fingers lose hold of the screwdriver, steering provides a larger action correction $\|\Delta a\|$, successfully correcting the grasp twice in the episode, and prevents the screwdriver from falling.}
\label{fig:qual_real}
\end{figure*}

\paragraph{Results}

On the continuous \emph{rotation} and \emph{palm-down reorientation} tasks, GUARD substantially improves real-robot performance over the unsteered BC policy (Table~\ref{tab:real_robot_summary}). For \emph{rotation}, the success rate improves by $40.0$ pp, and the drop rate reduces by $55.0$ pp. Fig.~\ref{fig:qual_real} shows how GUARD successfully steers the policy for the \emph{rotation} task. On \emph{palm-down reorientation}, GUARD improves the success rate by $25.0$ pp, converting those drops to successful reorientations. The terminal orientation error from the palm normal axis reduces from $4.4^\circ\!\to\!2.1^\circ$ and the longest stable object hold duration increases from $32.3\!\to\!37.5$\,s when evaluated on successful episodes for both methods, indicating tighter and more durable reorientation. GUARD also maintains a low RND score, comparable to BC, on successful episodes, indicating that steered rollouts remain in-distribution and do not encounter OOD states.
% In contrast, on the \emph{palm-up reorientation} task, GUARD almost entirely fails to steer. We believe this failure of GUARD is due to insufficient contact modeling in the simulation, as required in the real world. We noticed a significant screwdriver rotation mid-execution in the real world, which was infrequent under the same policy in simulation, resulting in a large sim-to-real gap. As a result, the Q-function was entirely unaware of such anomalous behavior, rendering its steering ineffective. We believe such gaps can be narrowed in future work using mixed buffers that capture real-world interactions, a technique that existing steering methods regularly employ, enabling the Q-function to learn the maximum possible real-world failures.
On the \emph{palm-up reorientation} task, GUARD hurts performance relative to BC. We attribute this to inaccurate contact modeling in simulation compared to the real world. We observed a pronounced screwdriver rotation mid-execution on the real robot that occurred rarely under the same policy in simulation, resulting in a large sim-to-real gap. 
%Because these transitions were essentially absent in the simulation for both expert RL and cloned BC policies, the anomaly metric would have been corrupted for them, preventing the Q-function from flagging this anomalous behavior and rendering its steering ineffective. 
Because these transitions were largely absent in the simulation, we hypothesize that the Q-function is unable to anticipate this behavior and thus fails to provide the required correction to recover to high-support regions.
We intend to resolve such gaps in future work by training with wider domain randomization and on mixed buffers that incorporate real-world interactions, a commonly used technique in steering methods for real-world deployment (LPB, DynaGuide). This can further close the sim-to-real gap, allowing the Q-function to be well aware of the failure modes actually encountered on hardware.

%% file: tables/adroit_state_4.tex
\definecolor{successbg}{RGB}{220,237,244}
\definecolor{logubg}{RGB}{253,235,221}
\definecolor{stdgray}{RGB}{105,105,105}
\definecolor{tasksep}{RGB}{170,170,170}

% Mean at normal size; standard deviation smaller and gray.
\newcommand{\meanstd}[2]{%
  \ensuremath{#1}\,
  {\color{stdgray}\tiny\ensuremath{\pm #2}}%
}

\begin{table*}[t]
\centering
\vspace{0.5em}
\caption{Adroit simulation results across nominal and perturbed dynamics. Here, $\sigma_d=0$ denotes nominal, while $\sigma_d>0$ denotes task-specific dynamics perturbations at the indicated noise level. Results are reported over three seeds using 100 test rollouts. \textbf{SR} denotes success rate ($\uparrow$), and \textbf{AS} denotes the RND-based state-anomaly score on successful episodes ($\downarrow$).}
\label{tab:adroit_state}
\renewcommand{\arraystretch}{1.12}

\resizebox{\textwidth}{!}{%
\begin{tabular}{
l c
!{\color{tasksep}\vrule width 0.3pt}
c c
c c
c c
c c
!{\color{tasksep}\vrule width 0.3pt}
c c
}
\toprule

&
&
\multicolumn{2}{!{\color{tasksep}\vrule width 0.3pt}c}{Relocate}
&
\multicolumn{2}{c}{Pen}
&
\multicolumn{2}{c}{Hammer}
&
\multicolumn{2}{c}{Door}
&
\multicolumn{2}{!{\color{tasksep}\vrule width 0.3pt}c}{Average}
\\

\cmidrule(lr){3-4}
\cmidrule(lr){5-6}
\cmidrule(lr){7-8}
\cmidrule(lr){9-10}
\cmidrule(lr){11-12}

Method
& Metric
& $\sigma_d = 0$ &  $\sigma_d = 0.1$
& $\sigma_d = 0$ &  $\sigma_d = 0.3$
& $\sigma_d = 0$ &  $\sigma_d = 0.1$
& $\sigma_d = 0$ &  $\sigma_d = 0.05$
& $\sigma_d = 0$ &  --
\\
\midrule

% ------------------------------------------------------------------
% BC
% ------------------------------------------------------------------
\rowcolor{successbg}
\cellcolor{white} BC
& SR
& \meanstd{0.74}{0.01}
& \meanstd{0.45}{0.06}
& \meanstd{0.75}{0.008}
& \meanstd{0.49}{0.04}
& \meanstd{0.52}{0.03}
& \meanstd{0.40}{0.05}
& \meanstd{0.37}{0.08}
& \meanstd{0.32}{0.05}
& \meanstd{0.60}{0.03}
& \meanstd{0.41}{0.05}
\\

\rowcolor{logubg}
\cellcolor{white}
& AS
& \meanstd{12.14}{4.76}
& \meanstd{\mathbf{25.34}}{4.51}
& \meanstd{16.81}{1.65}
& \meanstd{41.37}{18.62}
& \meanstd{105.16}{49.48}
& \meanstd{95.54}{38.36}
& \meanstd{19.94}{3.54}
& \meanstd{19.51}{1.17}
& \meanstd{38.51}{14.86}
& \meanstd{45.44}{15.67}
\\
\addlinespace[0.3em]

% ------------------------------------------------------------------
% Sparse-Q
% ------------------------------------------------------------------
\rowcolor{successbg}
\cellcolor{white} Sparse-Q
& SR
& \meanstd{0.80}{0.11}
& \meanstd{0.55}{0.13}
& \meanstd{0.88}{0.04}
& \meanstd{\mathbf{0.62}}{0.02}
& \meanstd{0.74}{0.09}
& \meanstd{0.66}{0.11}
& \meanstd{0.21}{0.07}
& \meanstd{0.19}{0.05}
& \meanstd{0.66}{0.08}
& \meanstd{0.50}{0.08}
\\

\rowcolor{logubg}
\cellcolor{white}
& AS
& \meanstd{12.67}{2.79}
& \meanstd{88.60}{28.70}
& \meanstd{5.71}{1.06}
& \meanstd{18.17}{1.06}
& \meanstd{58.58}{24.56}
& \meanstd{83.44}{19.47}
& \meanstd{24.23}{4.36}
& \meanstd{18.98}{2.34}
& \meanstd{25.30}{8.19}
& \meanstd{52.30}{12.89}
\\
\addlinespace[0.3em]

% ------------------------------------------------------------------
% LPB
% ------------------------------------------------------------------
\rowcolor{successbg}
\cellcolor{white} LPB
& SR
& \meanstd{0.78}{0.03}
& \meanstd{0.53}{0.03}
& \meanstd{0.74}{0.03}
& \meanstd{0.50}{0.03}
& \meanstd{0.57}{0.04}
& \meanstd{0.47}{0.03}
& \meanstd{0.49}{0.04}
& \meanstd{0.36}{0.03}
& \meanstd{0.64}{0.04}
& \meanstd{0.47}{0.03}
\\

\rowcolor{logubg}
\cellcolor{white}
& AS
& \meanstd{9.75}{3.28}
& \meanstd{42.95}{19.10}
& \meanstd{6.46}{2.18}
& \meanstd{49.17}{30.86}
& \meanstd{28.03}{10.13}
& \meanstd{39.07}{8.84}
& \meanstd{16.19}{2.13}
& \meanstd{16.93}{2.70}
& \meanstd{15.11}{4.43}
& \meanstd{37.03}{15.38}
\\
\addlinespace[0.3em]

% ------------------------------------------------------------------
% DynaGuide
% ------------------------------------------------------------------
\rowcolor{successbg}
\cellcolor{white} DynaGuide
& SR
& \meanstd{0.87}{0.03}
& \meanstd{0.55}{0.04}
& \meanstd{0.79}{0.03}
& \meanstd{0.54}{0.01}
& \meanstd{0.78}{0.04}
& \meanstd{0.79}{0.05}
& \meanstd{0.55}{0.02}
& \meanstd{0.51}{0.04}
& \meanstd{0.75}{0.03}
& \meanstd{0.60}{0.04}
\\

\rowcolor{logubg}
\cellcolor{white}
& AS
& \meanstd{9.18}{3.75}
& \meanstd{35.12}{5.39}
& \meanstd{21.68}{7.60}
& \meanstd{39.29}{10.62}
& \meanstd{22.01}{7.18}
& \meanstd{17.71}{7.58}
& \meanstd{17.37}{1.28}
& \meanstd{19.62}{0.26}
& \meanstd{17.56}{4.95}
& \meanstd{27.94}{5.96}
\\
\addlinespace[0.3em]

% ------------------------------------------------------------------
% UA-Flow
% ------------------------------------------------------------------
\rowcolor{successbg}
\cellcolor{white} UA-Flow
& SR
& \meanstd{0.75}{0.01}
& \meanstd{0.43}{0.02}
& \meanstd{0.73}{0.009}
& \meanstd{0.47}{0.03}
& \meanstd{0.51}{0.02}
& \meanstd{0.36}{0.009}
& \meanstd{0.39}{0.03}
& \meanstd{0.34}{0.05}
& \meanstd{0.60}{0.02}
& \meanstd{0.40}{0.03}
\\

\rowcolor{logubg}
\cellcolor{white}
& AS
& \meanstd{7.15}{1.37}
& \meanstd{46.10}{16.26}
& \meanstd{9.15}{2.85}
& \meanstd{38.42}{10.23}
& \meanstd{91.51}{18.02}
& \meanstd{84.32}{37.38}
& \meanstd{16.03}{0.94}
& \meanstd{20.30}{2.50}
& \meanstd{30.96}{5.80}
& \meanstd{47.29}{16.59}
\\
\addlinespace[0.3em]

% ------------------------------------------------------------------
% Joint-U
% ------------------------------------------------------------------
\rowcolor{successbg}
\cellcolor{white} U-SA
& SR
& \meanstd{0.84}{0.02}
& \meanstd{0.48}{0.02}
& \meanstd{0.75}{0.03}
& \meanstd{0.53}{0.02}
& \meanstd{0.77}{0.03}
& \meanstd{0.68}{0.05}
& \meanstd{0.52}{0.06}
& \meanstd{0.39}{0.05}
& \meanstd{0.72}{0.04}
& \meanstd{0.52}{0.04}
\\

\rowcolor{logubg}
\cellcolor{white}
& AS
& \meanstd{5.74}{1.02}
& \meanstd{45.15}{13.54}
& \meanstd{12.66}{6.21}
& \meanstd{35.84}{8.94}
& \meanstd{14.79}{5.37}
& \meanstd{32.25}{16.94}
& \meanstd{11.17}{0.49}
& \meanstd{17.40}{1.53}
& \meanstd{11.09}{3.27}
& \meanstd{32.66}{10.24}
\\

% Faint separator before our method
\arrayrulecolor{tasksep}
\specialrule{0.4pt}{0.35em}{0.35em}
\arrayrulecolor{black}

% ------------------------------------------------------------------
% GUARD
% ------------------------------------------------------------------
\rowcolor{successbg}
\cellcolor{white} \textbf{GUARD}
& SR
& \meanstd{\mathbf{0.98}}{0.005}
& \meanstd{\mathbf{0.79}}{0.03}
& \meanstd{\mathbf{0.93}}{0.01}
& \meanstd{0.60}{0.03}
& \meanstd{\mathbf{0.99}}{0.009}
& \meanstd{\mathbf{0.93}}{0.06}
& \meanstd{\mathbf{0.95}}{0.005}
& \meanstd{\mathbf{0.89}}{0.03}
& \meanstd{\mathbf{0.96}}{0.008}
& \meanstd{\mathbf{0.80}}{0.04}
\\

\rowcolor{logubg}
\cellcolor{white}
& AS
& \meanstd{\mathbf{4.21}}{1.89}
& \meanstd{62.49}{31.94}
& \meanstd{\mathbf{4.04}}{0.53}
& \meanstd{\mathbf{16.83}}{1.89}
& \meanstd{\mathbf{1.19}}{0.66}
& \meanstd{\mathbf{3.46}}{1.72}
& \meanstd{\mathbf{3.29}}{0.80}
& \meanstd{\mathbf{3.92}}{0.60}
& \meanstd{\mathbf{3.18}}{0.97}
& \meanstd{\mathbf{21.68}}{9.04}
\\

\bottomrule
\end{tabular}%
}
\end{table*}

%% file: tables/anomaly_ablation.tex
% \begin{table}[htbp]
% \centering
% \small
% \setlength{\tabcolsep}{6pt}
% \begin{tabular}{cccccc}
% \toprule
% logpZO & RND & Recon-AE & Dyn-Err & Dyn-Ens-Var \\
% \midrule
% $\mathbf{0.95 \pm 0.005}$ & $0.89 \pm 0.03$ & $0.90 \pm 0.02$ & $0.61 \pm 0.12$ & $0.91 \pm 0.008$ \\
% \bottomrule
% \end{tabular}
% \caption{Anomaly-Signal Ablation.}
% \label{tab:anomaly_ablation_door}
% \end{table}

\begin{table}[htbp]
\centering
\footnotesize
\setlength{\tabcolsep}{2.5pt}
\renewcommand{\arraystretch}{0.95}
\begin{tabular}{ccccc}
\toprule
logpZO & RND & Recon-AE & Dyn-Err & Dyn-Ens \\
\midrule
$\mathbf{.95{\pm}.005}$ &
$.89{\pm}.03$ &
$.90{\pm}.02$ &
$.61{\pm}.12$ &
$.91{\pm}.008$ \\
\bottomrule
\end{tabular}
\vspace{-0.25em}
\caption{Anomaly-Signal Ablation on \textit{Door} (SR $\uparrow$).}
\label{tab:anomaly_ablation_door}
\vspace{-1em}
\end{table}

% \begin{wraptable}{r}{0.34\textwidth}
% \centering
% \small
% \vspace{-\baselineskip}
% \setlength{\tabcolsep}{6pt}
% \begin{tabular}{lc}
% \toprule
% Anomaly signal & Success \\
% \midrule
% logpZO & $\mathbf{0.953 \pm 0.005}$ \\
% Dyn-Ens-Var   & $0.910 \pm 0.008$ \\
% Recon-AE      & $0.897 \pm 0.021$ \\
% RND           & $0.893 \pm 0.034$ \\
% Dyn-Err       & $0.613 \pm 0.119$ \\
% \bottomrule
% \end{tabular}
% \caption{Anomaly-Signal Ablation.}
% \label{tab:anomaly_ablation_door}
% \end{wraptable}

%% file: tables/real_world.tex
\begin{table}[htbp]
\centering
\caption{Real-robot Allegro V5 Plus screwdriver manipulation results for BC and GUARD, with $n=20$ trials per method. The best result for each metric is shown in \textbf{bold}.}
\label{tab:real_robot_summary}

\setlength{\tabcolsep}{3pt}
\renewcommand{\arraystretch}{1.08}

\resizebox{\columnwidth}{!}{%
\begin{tabular}{@{}lcccccc@{}}
\toprule
&
\multicolumn{2}{c}{Success (\%) $\uparrow$}
&
\multicolumn{2}{c}{Drop (\%) $\downarrow$}
&
\multicolumn{2}{c}{RND $\downarrow$}
\\

\cmidrule(lr){2-3}
\cmidrule(lr){4-5}
\cmidrule(lr){6-7}

Task
& BC
& GUARD
& BC
& GUARD
& BC
& GUARD
\\
\midrule

Rotation
& 20
& \textbf{60}
& 80
& \textbf{25}
& 0.11\,{\scriptsize\textcolor{gray}{$\pm$0.03}}
& 0.07\,{\scriptsize\textcolor{gray}{$\pm$0.02}}
\\

Palm-Down
& 70
& \textbf{95}
& 30
& \textbf{5}
& 0.05\,{\scriptsize\textcolor{gray}{$\pm$0.11}}
& 0.03\,{\scriptsize\textcolor{gray}{$\pm$0.07}}
\\

Palm-Up
& \textbf{30}
& 10
& \textbf{5}
& 20
& 0.01\,{\scriptsize\textcolor{gray}{$\pm$0.00}}
& 0.02\,{\scriptsize\textcolor{gray}{$\pm$0.01}}
\\

\bottomrule
\end{tabular}%
}
\end{table}

%% file: sections/7_conclusion.tex
\section{Conclusion and Future Work}
\label{sec:conclusion}

% We introduced \textbf{GUARD}, an anomaly-informed Q-guidance framework for steering frozen generative robot policies at inference time. GUARD addresses a key limitation of Q-guided denoising in robotics: sparse terminal task rewards provide limited intermediate supervision for learning informative Q-functions. By augmenting task success with dense anomaly signals learned from the base policy's own training distribution, GUARD learns long-horizon values that capture both task completion and distributional support. These values are used to steer the denoising process toward locally higher-return actions while keeping the system closed to the high-support manifold.
We introduced \textbf{GUARD}, an anomaly-informed Q-guidance framework for robustly steering frozen generative robot policies. Rather than relying on task-specific dense rewards or manual failure annotations, GUARD leverages anomaly signals learned from the base policy's training distribution and combines them with sparse task rewards to learn Q-functions. By propagating this information through the Bellman return, GUARD enables the value gradients to account for both task completion and the likelihood of entering poorly supported states, rather than greedily minimizing instantaneous uncertainty.

Across Adroit dexterous manipulation tasks, GUARD significantly improves task success while maintaining lower anomaly scores than existing steering methods, including under perturbed dynamics. We further demonstrate direct sim-to-real transfer without any real-robot training, achieving performance gains on two of three evaluated tasks. Together, these results demonstrate that signals typically used for runtime failure monitoring can also provide scalable supervision to actively improve generative robot policies through test-time Q-steering methods.

Our results indicate that our anomaly-informed Q-steering can transfer surprisingly well across shifts in dynamics and even to the real world. We hypothesize that task conditions with overlapping successful state distributions and comparable dynamics can reliably transfer the anomaly metrics and subsequent Q-functions, as demonstrated by successful steering across perturbed simulation dynamics and by rotation and palm-down reorientation results in the real world. However, if the environment shift significantly alters the successful state distribution, turning nominal states anomalous and leaving the Q-function unaware of real failure modes, GUARD may no longer provide the required correction. This is exemplified by the palm-up reorientation results.

Future work will explore resolving these issues through mixed simulation and real-world training to further improve real-world transfer and explore extensions to policies with multimodal inputs and broader manipulation tasks.